\documentclass[10pt,twocolumn,letterpaper]{article}

\usepackage{iccv}
\usepackage{times}
\usepackage{epsfig}
\usepackage{graphicx}
\usepackage{amsmath}
\usepackage{amssymb}
\usepackage{booktabs}
\usepackage{parskip}

\usepackage[breaklinks=true,bookmarks=false]{hyperref}

\iccvfinalcopy 

\def\iccvPaperID{17} 

\ificcvfinal\fi

\newcommand{\partners}[0]{American Red Cross}
\newcommand{\partnersabbrv}[0]{ARC}

\begin{document}

\title{Rapid building damage assessment workflow: An implementation for the 2023 Rolling Fork, Mississippi tornado event}

\author{
Caleb Robinson\textsuperscript{1}\thanks{Corresponding author: \texttt{caleb.robinson@microsoft.com}}, \enskip Simone Fobi Nsutezo\textsuperscript{1}, \enskip Anthony Ortiz\textsuperscript{1}, \enskip Tina Sederholm\textsuperscript{1}, \enskip Rahul Dodhia\textsuperscript{1},\\ Cameron Birge\textsuperscript{2}, \enskip Kasie Richards\textsuperscript{3}, \enskip Kris Pitcher\textsuperscript{3}, \enskip Paulo Duarte\textsuperscript{3}, \enskip Juan M. Lavista Ferres\textsuperscript{1} \\ \\
    \textsuperscript{1}Microsoft AI for Good Research Lab, \enskip \textsuperscript{2}Microsoft Philanthropies, \\
    \textsuperscript{3}American Red Cross
}

\maketitle
\ificcvfinal\thispagestyle{empty}\fi

\begin{abstract}
Rapid and accurate building damage assessments from high-resolution satellite imagery following a natural disaster is essential to inform and optimize first responder efforts. However, performing such building damage assessments in an automated manner is non-trivial due to the challenges posed by variations in disaster-specific damage, diversity in satellite imagery, and the dearth of extensive, labeled datasets. To circumvent these issues, this paper introduces a human-in-the-loop workflow for rapidly training building damage assessment models after a natural disaster. This article details a case study using this workflow, executed in partnership with the \partners{} during a tornado event in Rolling Fork, Mississippi in March, 2023. The output from our human-in-the-loop modeling process achieved a precision of 0.86 and recall of 0.80 for damaged buildings when compared to ground truth data collected post-disaster. This workflow was implemented end-to-end in under 2 hours per satellite imagery scene, highlighting its potential for real-time deployment.
\end{abstract}

\begin{figure*}[t]
    \centering
    \includegraphics[width=0.49\linewidth]{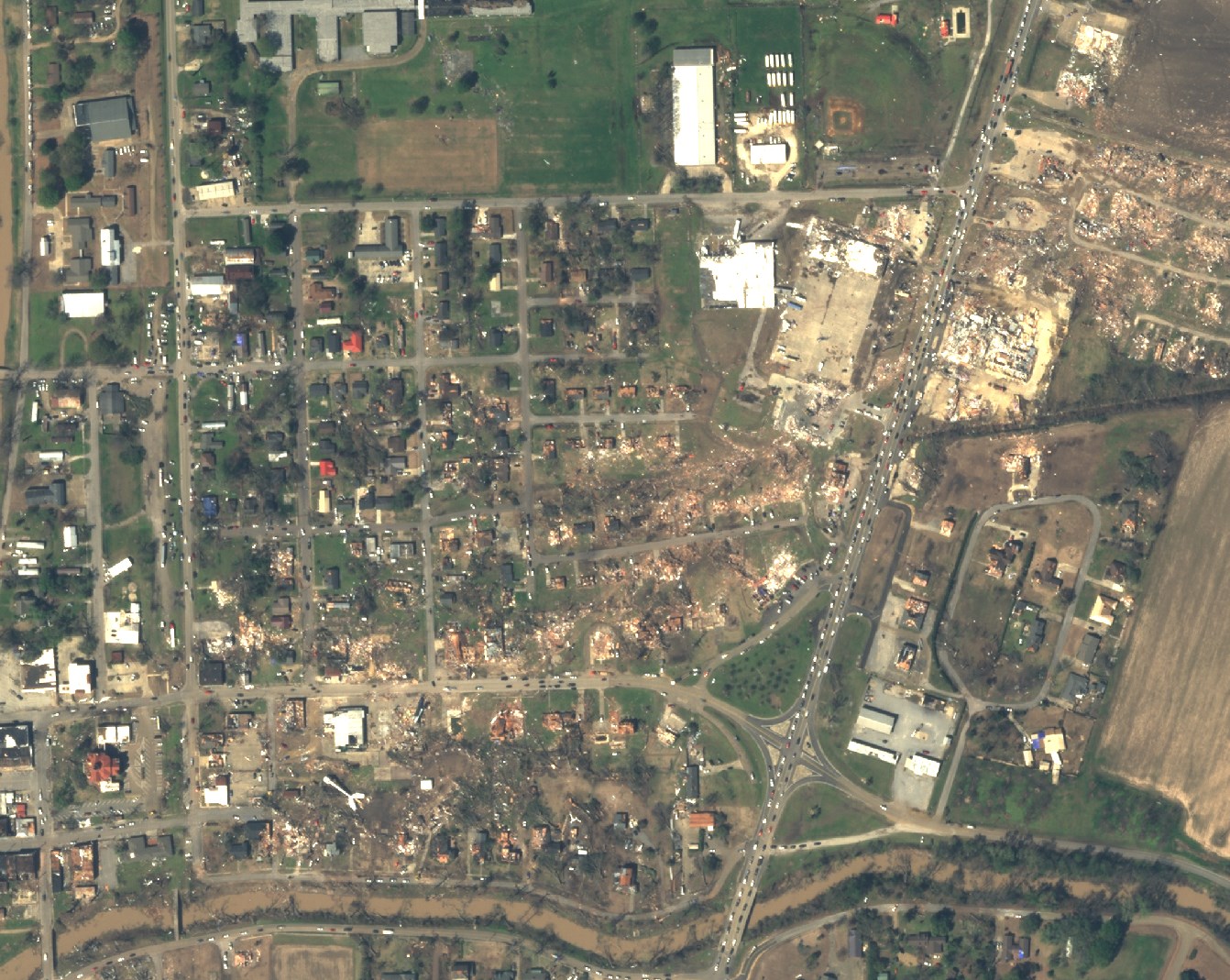}
    \includegraphics[width=0.49\linewidth]{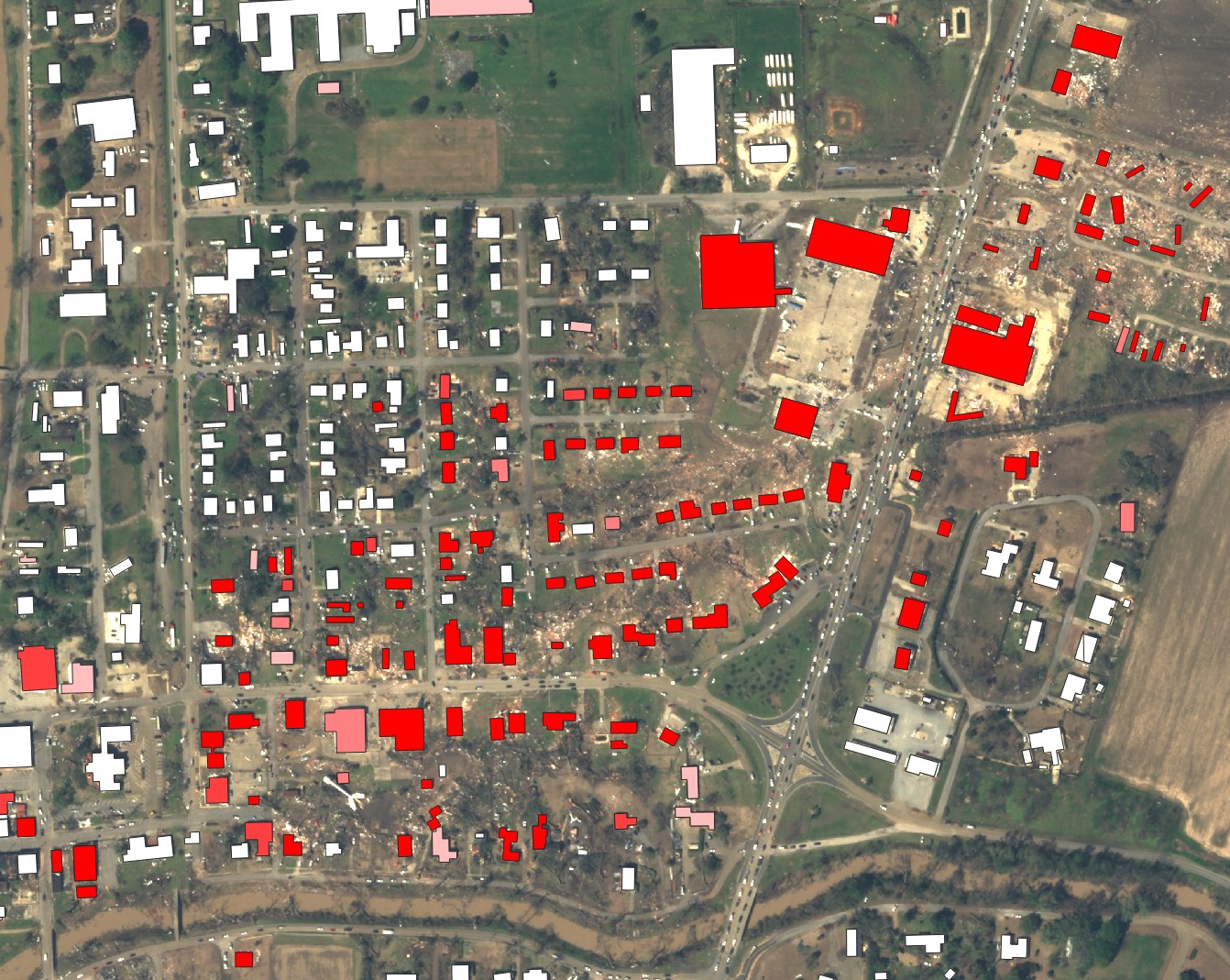}
    \caption{(\textbf{Left}) Post disaster Planet SkySat imagery over Rolling Fork, Mississippi from March 25th; (\textbf{Right}) Building level model damage estimates (color gradient indicates damage from no damage colored in white to destroyed in red).}
    \label{fig:post}
\end{figure*}

\section{Introduction}
Rapid building damage assessment in the wake of a natural disaster can inform first responders about the types and magnitudes of resources needed on the ground to support affected populations. Advancements in automated building damage assessment have been supported by the availability of high resolution imagery pre and post a disaster paired with corresponding damage labels in datasets such as xBD~\cite{gupta2019xbd} and xFBD~\cite{melamed2022xfbd}.
Variants of semantic segmentation methods have been applied to pre and post-disaster imagery to perform the dual tasks of outlining building footprints and classify damaged buildings (\cite{gupta2021rescuenet,khvedchenya2021fully,bai2020pyramid,gholami2022deployment}). Despite these advancements, building damage assessment models suffer from significant degradation when evaluated on out-of-distribution datasets~\cite{benson2020assessing,gholami2022deployment} due to the breadth of variations in the type of disaster, image modality, capture conditions, weather, and data size, among others. These issues limit the usability of existing disaster assessment models in real world scenarios where quick and accurate results are needed for a specific area using any available imagery.

We present a rapid response damage assessment workflow that utilizes a small number of labels and only post-disaster imagery. We demonstrate a fast mechanism by which a model can be tuned to perform well under a new disaster using only a small number of training examples thereby eliminating the need for generalization. In addition, this method reduces the dependence on pre-post image pairs, resulting in a more streamlined damage assessment pipeline. This workflow was deployed immediately after the occurrence of a natural disaster and validated in collaboration with the \partners{} (\partnersabbrv), showing that this approach can support rapid disaster response.

\section{Application Context}
On \textbf{Friday, March 24, 2023,} an EF4 tornado formed in Mississippi. The tornado lasted for 71 minutes and caused 16 fatalities and 165 injuries. It first touched down in Issaquena County, and tracked northeastward through Sharkey and Humphreys Counties, causing extensive damage in Rolling Fork, Midnight, and Silver City. The most severe damage was observed in Rolling Fork, where numerous structures were destroyed or severely damaged, including homes, businesses, a hospital, schools, and a water tower. Throughout its path, the tornado produced varying degrees of damage, from minor tree and power pole damage to catastrophic destruction of buildings and infrastructure. The tornado eventually dissipated after having traveled 59.4 miles (95.6 km) in total.

On \textbf{March 25th}, our partners at the \partnersabbrv{} suggested that we use the event to test our damage assessment workflow, to provide them with satellite imagery-based damage analysis to inform ground operations that would occur over the following week. The same day, a Planet Labs SkySat satellite captured a clear post-disaster image of the town of Rolling Fork, which provided us with the necessary data to implement our workflow (described in Section \ref{sec:workflow}).

On \textbf{March 26th}, we executed our damage assessment workflow, and provided the \partnersabbrv{} with a set of 1347 building footprints (from the Global Microsoft Building ML Footprints dataset~\cite{buildingFootprints}), and an estimate of how damaged each building was on a scale of 0 to 100. 

Over the following weeks, the \partnersabbrv{} deployed more than 800 trained disaster workers to respond to the event. These disaster workers provided sheltering to those displaced by the disaster as well as supported health, mental health services, and recovery support. \partnersabbrv{} disaster workers provided shelf-stable meals in the community, where other resources were not available and provided relief items including comfort kits to people in need. The disaster response team trained in damage assessment conducted more than 3500 assessments of residential damage across the state of Mississippi. The ground truth damage assessment included over 909 addresses in Rolling Fork where they categorized the building at each address according to the FEMA building damage scale: no visible damage, affected, minor damage, major damage, or destroyed~\cite{fema2021}.

In the following sections we outline our damage assessment workflow, describe our modeled output over Rolling Fork (as shown in Figure \ref{fig:post}), and perform a post-hoc analysis of the model output using the ground truth data.

\section{Damage Assessment Workflow} \label{sec:workflow}
Given a new disaster area of interest (\textit{AOI}),  we outline our building damage assessment workflow:

\paragraph{Satellite image and building footprint acquisition.} We acquire \textit{post-disaster} high-resolution satellite imagery \textit{scenes} overlapping the AOI. A \textit{scene} is defined as an image captured from the same pass by a satellite\footnote{Large AOIs are often made up of multiple scenes from potentially different satellite passes at different times of day, with different imaging angles, cloud coverage, illumination, and on-the-ground conditions.}. We obtain corresponding building footprints (polygons) for the same scene from OpenStreetMap~\cite{OpenStreetMap} or the Microsoft Building Footprint dataset~\cite{buildingFootprints}. 

\paragraph{Satellite image-based label acquisition.} We set up an instance of the ``satellite imagery labeling tool''~\cite{satelliteImageryLabelingTool} with post-disaster imagery. We manually label a random selection of building footprints in the AOI as ``damaged'' or ``not damaged'' based on a visual inspection of the post-disaster imagery and use them for validation of the modeling output. 

\paragraph{Model training.} First, we manually label a total of $\approx 100$ example polygons of ``background'', ``building'', and ``damaged building'' classes using the labeling tool instance. Next, we fine-tune an pre-trained semantic segmentation model \footnote{We used a U-Net with an ImageNet pre-trained ResNet-50 backbone as our semantic segmentation architecture, but practitioners can choose alternative architectures with pre-trained models better suited to the task.} with the cumulative labeled examples. We run the fine-tuned model on the entire scene to classify each pixel as: ``background'', ``building'', or ``damaged building''. We compute, for each building footprint in the AOI, the percentage area of the footprint that is classified as ``damaged building'', $\hat{y}_i$. These damage proportions are used to represent the likelihood that a building is damaged.
This process is repeated, per scene, until the performance of the modeled output is acceptable (with respect to the validation labels).

Formally, the output of this building damage assessment process is a set of $N$ building footprints (polygons) and the associated damage estimates, $\{\hat{y}_i\}_{i=1}^N$ where $\hat{y}_i$ is the estimated percent damage of the $i^{\text{th}}$ building, i.e. $\hat{y}_i \in \left[0,100\right]$. 

\paragraph{Validation.}
\label{subsec:validation}
Given building level percent damage estimates obtained from a model, we determine the performance of the model by computing a set of validation metrics with respect to labeled building level data. Here, labeled building level data can be obtained either from a ground truth assessment by experts (e.g. a categorization of each building according to the FEMA building damage criteria) or, less desirably, by an independent visual interpretation of the post-disaster imagery. Formally, we assume that we have $N$ labels over $C$ damage classes, $\{y_i\}_{i=1}^N$, where $y_i \in \{1, \cdots, C\}$  and the damage classes represent categories such as ``no visible damage'', ``affected'', ``minor damage'', ``major damage'', and ``destroyed''.

First, we aggregate the $C$ damage classes into two classes, i.e. converting the problem into a binary classification problem. For example, if our damage labels follow the FEMA categorization, we may consider ``major damage'' and ``destroyed'' as a single ``damaged'' class with the remaining categories grouped into a ``not damaged'' class.

Now, given a percent damage estimate, $\hat{y}_i$, and binary damage label, $y_i$, for each building, we compute precision and recall for the damaged class under an \textit{any damage} threshold (i.e. where $\hat{y}_i > 0$), as well as a precision-recall curve and average precision metric. More specifically, given a threshold value, $\theta$, we compute the following:

\begin{description}
\item[$\text{tp}_\theta$] \textit{true positives}, the number of buildings in which $\hat{y}_i > \theta$ and $y_i = 1$, i.e. the number of buildings where our model estimated a damage percentage greater than $\theta$ and the building was ``damaged''.
\item[$\text{fp}_\theta$] \textit{false positives}, the number of buildings in which $\hat{y}_i > \theta$ and $y_i = 0$, i.e. the number of buildings where our model estimated a damage percentage greater than $\theta$ and the building was ``not damaged''.
\item[$\text{fn}_\theta$] \textit{false negatives}, the number of buildings in which $\hat{y}_i \leq \theta$ and $y_i = 1$, i.e. the number of buildings where our model estimated a damage percentage less than or equal to $\theta$ and the building was ``damaged''.
\end{description}

Precision and recall under the any damage threshold ($\theta$ = 0) are then:
\begin{align}
    \text{\textbf{Precision}}_0 &= \frac{\text{tp}_0}{\text{tp}_0 + \text{fp}_0}\\
    \text{\textbf{Recall}}_0&= \frac{\text{tp}_0}{\text{tp}_0 + \text{fn}_0}
\end{align}

Intuitively, precision is the fraction of buildings that the model estimates to be damaged, that are, in fact, damaged. E.g. if precision is 0.8, then 1 out of every 5 buildings the model predicts to be damaged, will not be damaged. Similarly, recall is the fraction of damaged buildings that the model correctly estimates to be damaged. E.g. if recall is 0.8 then the model identified 80\% of the damaged buildings in the scene.

\begin{figure}[b]
    \centering
    \includegraphics[width=1.0\linewidth]{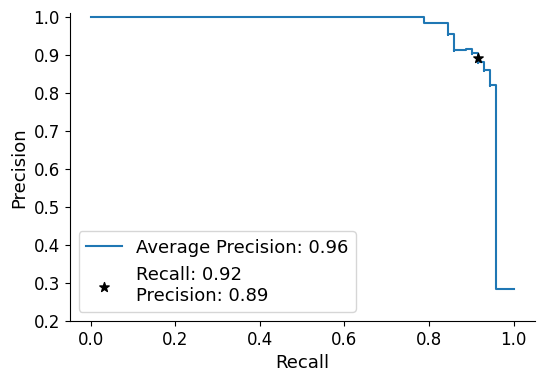}
    \caption{Precision-recall curve for our building damage estimates compared to image interpretation based labels.}
    \label{fig:prcurve1}
\end{figure}

Setting $\theta$ to all possible values between 0 and 1 and plotting the resulting precision and recall values gives us a ``precision-recall curve'' that shows the trade off between precision and recall. As the percent damage threshold with which we use to classify a building as damaged increases, the precision of the model will increase while the recall decreases. Intuitively, the precision-recall curve serves as proxy for damage localization, where ``major damage'' and ``destroyed'' buildings should have large percent damage estimates.

Finally, \textbf{average precision} (AP) is a summary of precision and recall across multiple thresholds. Specifically, AP is a weighted mean of precisions over the set of possible $\theta$ values, where the weighting is obtained from the change in recall between thresholds~\cite{scikit-learn}:
\begin{equation}
\text{\textbf{AP}} = \sum_n {(\text{Recall}_{\theta_n} - \text{Recall}_{\theta_{n-1}}) * \text{Precision}_{\theta_n}} 
\end{equation} 

\section{Rolling Fork, Mississippi Results}

\begin{figure*}[t]
    \centering
    \includegraphics[width=0.49\linewidth]{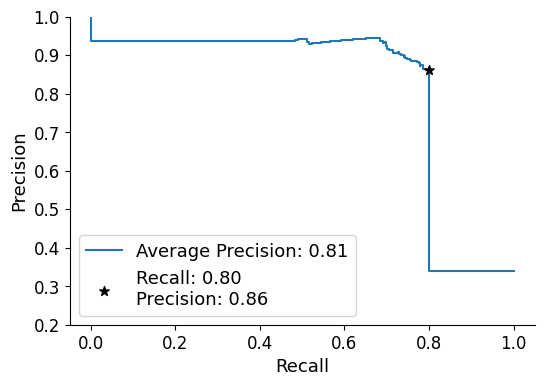}
    \includegraphics[width=0.49\linewidth]{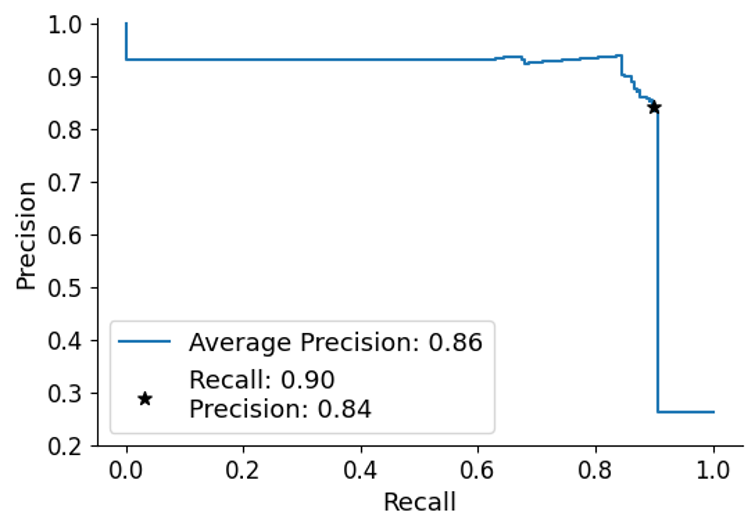}
    \caption{Precision-recall curves for our building damage estimates compared to ground truth labels under two definitions of ``damaged'' buildings. The (\textbf{left}) panel shows the result when ``major damage'' and ``destroyed'' points are considered ``damaged'' while the (\textbf{right}) panel shows the result when only the ``destroyed'' points are considered ``damaged''.}
    \label{fig:prcurve}
\end{figure*}

For our Rolling Fork building damage assessment, we produced percent damage estimates for 1,347 buildings over the city of Rolling Fork, Mississippi using Planet SkySat imagery from March 25th, 2023\footnote{Scene ID: 20230325\_195842\_ssc11\_u0001.}. We annotated 153 examples of ``background'' (34 examples), ``undamaged building'' (71 examples), and ``damaged building'' (48 examples), then used these to train a semantic segmentation model that can make predictions over the entire image.

We perform two validation efforts -- using labeled data collected through a visual interpretation of the satellite imagery and using labeled data collected through a ground truth analysis. We detail the results of these efforts below.

\subsection{Validation with image-based labels}
Our initial validation effort involves annotating a set of 250 randomly selected building footprints as ``damaged'' or ``not damaged'' based on an interpretation of the post-disaster imagery that was independent of the annotation effort used to train the damage assessment model. Our building level damage estimates have an average precision of 0.96 with a precision-recall curve shown in Figure \ref{fig:prcurve1}. The precision and recall under the ``any damage'' threshold are 0.96 and 0.74.

\subsection{Validation with ground truth based labels}
We received 909 labels from the \partnersabbrv{} based on a ground truth analysis that they performed during their disaster response. These labels were assigned to \textit{addresses} and categorized according to the FEMA building damage assessment scale. We first geocoded the addresses to obtain latitude, longitude point locations, and matched them to the set of building footprints that we performed the damage assessment over. Due to noise in the geocoding process and the fact that not all buildings were analyzed by the \partnersabbrv{}, sometimes the ground truth points are not contained in a building footprint\footnote{Examples are i) the point occurs along a road, ii) several points occur on a single building or iii) a building has no matching points.}. Thus, we associate each ground truth point with the nearest building footprint in a 20 meter radius (if any).

We compute the validation metrics over the 830 ground truth points that are associated with a building footprint.

If we consider ground truth points that are categorized as ``major damage'' or ``destroyed'' to be the ``damaged'' class, and the remaining categories to be an ``undamaged'' class, precision and recall under the ``any damage'' threshold are 0.862 and 0.801. When only the ``destroyed'' class is considered to be ``damaged'', then precision and recall are 0.84 and 0.90. Figure \ref{fig:prcurve} shows the localization performance of the model where average precision increases from 0.81 (``major'' and ``destroyed'') to 0.86 (only ``destroyed'').

\section{Conclusion}
Our work demonstrates that a human-in-the-loop workflow can provide a rapid and effective method of building damage assessment post-natural disasters. Our method has demonstrated its practical usefulness by quickly training with minimal data, effectively bypassing the requirement for extensive labeled datasets and the challenges of training models that account for diverse disaster scenarios and satellite imagery variations. A key advantage of the workflow is its capability to be implemented in under 2 hours per satellite imagery scene by any user familiar with annotating satellite imagery, which enables its potential use in production scenarios where immediate action and response are of paramount importance. This was evident in the case of the tornado event in Rolling Fork, Mississippi, where the model, having been trained and validated collaboratively with the \partnersabbrv{}, resulted in a precision of 0.86 and a recall of 0.80 when evaluated against ground truth data. Future work will focus on refining this workflow, exploring scalability across disaster types, and strengthening collaboration between AI and human expertise.

\section*{Acknowledgments}
This work builds off of previous work done between Microsoft Philanthropies and the Netherlands Red Cross 510 data team. We would like to thank the anonymous reviewers for their feedback.
 
{\small
\bibliographystyle{ieee_fullname}
\bibliography{citations}
}

\end{document}